\PassOptionsToPackage{unicode}{hyperref}
\PassOptionsToPackage{hyphens}{url}
\PassOptionsToPackage{dvipsnames,svgnames,x11names}{xcolor}
\documentclass[
  11pt,
]{article}
\usepackage{amsmath,amssymb}
\usepackage{iftex}
\ifPDFTeX
  \usepackage[T1]{fontenc}
  \usepackage[utf8]{inputenc}
  \usepackage{textcomp} 
\else 
  \usepackage{unicode-math} 
  \defaultfontfeatures{Scale=MatchLowercase}
  \defaultfontfeatures[\rmfamily]{Ligatures=TeX,Scale=1}
\fi
\usepackage{lmodern}
\ifPDFTeX\else
\fi
\IfFileExists{upquote.sty}{\usepackage{upquote}}{}
\IfFileExists{microtype.sty}{
  \usepackage[]{microtype}
  \UseMicrotypeSet[protrusion]{basicmath} 
}{}
\usepackage{xcolor}
\usepackage[margin=1in]{geometry}
\usepackage{graphicx}
\makeatletter
\def\maxwidth{\ifdim\Gin@nat@width>\linewidth\linewidth\else\Gin@nat@width\fi}
\def\maxheight{\ifdim\Gin@nat@height>\textheight\textheight\else\Gin@nat@height\fi}
\makeatother
\setkeys{Gin}{width=\maxwidth,height=\maxheight,keepaspectratio}
\makeatletter
\def\fps@figure{htbp}
\makeatother
\newlength{\cslhangindent}
\newlength{\csllabelwidth}
\newlength{\cslentryspacingunit} 
\newenvironment{CSLReferences}[2] 
 {
  \setlength{\parindent}{0pt}
  \ifodd #1
  \let\oldpar\par
  \def\par{\hangindent=\cslhangindent\oldpar}
  \fi
  \setlength{\parskip}{#2\cslentryspacingunit}
 }%
 {}
\usepackage{calc}

\usepackage{accents}
\usepackage{amsmath}
\usepackage{bm}
\usepackage{booktabs}
\usepackage{braket}
\usepackage{caption}
\usepackage{dcolumn}
\usepackage{epigraph}
\usepackage{extarrows}
\usepackage{float}
\usepackage[bottom,flushmargin,hang,multiple]{footmisc}
\usepackage{graphicx}
\usepackage{hyperref}
\usepackage[export]{adjustbox}
\usepackage{libertine}
\usepackage[libertine]{newtxmath}
\usepackage[mathcal]{euscript}
\usepackage{mathrsfs}
\usepackage{multirow}
\usepackage{quoting}
\usepackage{scrextend}
\usepackage{setspace}
\usepackage{soul}
\usepackage{subcaption}
\usepackage{tabularx}
\usepackage{titlesec}
\usepackage[utf8]{inputenc}
\ifLuaTeX
  \usepackage{selnolig}  
\fi
\IfFileExists{bookmark.sty}{\usepackage{bookmark}}{\usepackage{hyperref}}
\IfFileExists{xurl.sty}{\usepackage{xurl}}{} 
\hypersetup{
  pdftitle={Silico-centric Theory of Mind},
  colorlinks=true,
  linkcolor={Maroon},
  filecolor={Maroon},
  citecolor={Blue},
  urlcolor={blue},
  pdfcreator={LaTeX via pandoc}}

\title{Silico-centric Theory of Mind}
\author{}
\date{\vspace{-2.5em}}

\begin{document}
\maketitle

\begin{center}
\author
{Anirban Mukherjee,$^{1\ast}$ Hannah H. Chang$^{2}$\\
\medskip
\normalsize{$^{1}$Samuel Curtis Johnson Graduate School of Management, Cornell University,}\\
\normalsize{Sage Hall, Ithaca, NY 14850, USA}\\
\normalsize{$^{2}$Lee Kong Chian School of Business, Singapore Management University,}\\
\normalsize{50 Stamford Road, Singapore, 178899}\\
\smallskip
\normalsize{$^\ast$To whom correspondence should be addressed; E-mail: am253\@cornell.edu.}\\
}
\end{center}
\medskip

\doublespacing

\quotingsetup{indentfirst=false, leftmargin=2em, rightmargin=2em, vskip=1ex}
\singlespacing

\begin{center} 
\noindent \textbf{Abstract}
\end{center}

\noindent Theory of Mind (ToM) refers to the ability to attribute mental
states, such as beliefs, desires, intentions, and knowledge, to oneself
and others, and to understand that these mental states can differ from
one's own and from reality. We investigate ToM in environments with
multiple, distinct, independent AI agents, each possessing unique
internal states, information, and objectives. Inspired by human
false-belief experiments, we present an AI (`focal AI') with a scenario
where its clone undergoes a human-centric ToM assessment. We prompt the
focal AI to assess whether its clone would benefit from additional
instructions. Concurrently, we give its clones the ToM assessment, both
with and without the instructions, thereby engaging the focal AI in
higher-order counterfactual reasoning akin to human mentalizing--with
respect to humans in one test and to other AI in another.

We uncover a discrepancy: Contemporary AI demonstrates near-perfect
accuracy on human-centric ToM assessments. Since information embedded in
one AI is identically embedded in its clone, additional instructions are
redundant. Yet, we observe AI crafting elaborate instructions for their
clones, erroneously anticipating a need for assistance. An independent
referee AI agrees with these unsupported expectations. Neither the focal
AI nor the referee demonstrates ToM in our `silico-centric' test.

\begin{center}\rule{0.5\linewidth}{0.5pt}\end{center}

\noindent Keywords: Theory of Mind, Counterfactual Reasoning, Artificial
Intelligence, Multi Entity AI Systems.

\newpage
\doublespacing

\hypertarget{introduction}{%
\section{Introduction}\label{introduction}}

Theory of Mind (ToM) is a fundamental cognitive ability that enables
individuals to attribute and comprehend mental states---such as beliefs,
intentions, desires, and emotions---both in themselves and in others
(\protect\hyperlink{ref-guajardo2016contribution}{Guajardo and
Cartwright 2016}, \protect\hyperlink{ref-premack1978does}{Premack and
Woodruff 1978}). This capability is foundational to our understanding
that others may possess perspectives and desires distinct from our own.
It plays a pivotal role in decision-making, empathy, and moral
reasoning, facilitating nuanced social interactions and communication.

A critical aspect of ToM is counterfactual thinking, a cognitive process
that involves imagining alternative realities by posing `what if'
questions (\protect\hyperlink{ref-mandel2007psychology}{Mandel et al.
2007}, \protect\hyperlink{ref-roese1997counterfactual}{Roese 1997}).
This type of thinking enables individuals to consider outcomes different
from those that have actually occurred, allowing for a deeper
understanding of potential consequences and the exploration of various
scenarios. While closely related to causal reasoning (e.g., imagining a
chain of elements like `woodpecker'-`pecks'-`hole'), counterfactual
thinking is distinguished by the inclusion of imagining the consequences
of environmental changes (`softwood' vs.~`hardwood', `rainforest'
vs.~`desert') such that a causal sequence may vary (e.g., `if the tree
were made of softwood, the woodpecker could peck through it more
easily') or be unlikely to occur (e.g., `conceiving of the Sahara desert
with few trees and no woodpeckers'). A counterfactual may exist only to
aid in causal reasoning but may also extend far beyond it, including
situations and circumstances that are distinct from the original.

The question of whether Artificial Intelligence (AI) can demonstrate ToM
has become a focal point of contemporary research. Emerging studies
present a mixed picture: some suggest that modern connectionist AI
systems exhibit nascent forms of ToM, akin to those observed in young
human children (e.g., \protect\hyperlink{ref-trott2023large}{Trott et
al. 2023}). Others highlight AIs' proficiency in causal reasoning tasks,
such as solving Raven's progressive matrices, which assess abstract
reasoning skills (\protect\hyperlink{ref-webb2023emergent}{Webb et al.
2023}), and conducting property induction tasks that evaluate inductive
reasoning (\protect\hyperlink{ref-han2024inductive}{Han et al. 2024}).
Research by Huang et al. (\protect\hyperlink{ref-huang2023clomo}{2023})
has shown AIs' capability for logical counterfactual thinking, though it
still falls short of human proficiency.

However, other research indicates that manifestations of such abilities
rely more on lexical cues rather than a deep understanding of mental
states or alternative realities. For instance, Ullman
(\protect\hyperlink{ref-ullman2023large}{2023}) finds that AIs struggle
when minor linguistic alterations to standard tests are introduced. Such
findings suggest AIs lack an intuitive understanding of mental states
and complex reasoning, instead relying on formulaic regurgitations of
memorized responses. These arguments align with theoretical critiques
that limitations inherent to AI preclude true counterfactual reasoning
(\protect\hyperlink{ref-chomsky2023noam}{Chomsky et al. 2023},
\protect\hyperlink{ref-pearl2018ai}{Pearl and Mackenzie 2018}). They
also resonate with ideas such as those of Lake et al.
(\protect\hyperlink{ref-lake2017building}{2017}) and Rabinowitz et al.
(\protect\hyperlink{ref-rabinowitz2018machine}{2018}), who suggest
desiderata and propose alternate approaches to model building whereby
deeper reasoning capacities may emerge.

Crucially, these investigations focus on human-centric scenarios, where
AIs are asked about the mental states of humans whose thoughts and
behaviors are described in vignettes. Similar to how the design of AI,
vis-à-vis neural networks, traces its roots to the biological origins of
cognition---the neural pathways that make up the human
mind---examinations of macro phenomena in machine cognition take
inspiration from the study of human psychology, through the
interpretation and implementation of classic and traditional assessment
methods and techniques for AI (e.g.,
\protect\hyperlink{ref-firestone2020performance}{Firestone 2020}).
Consequently, it is only natural for the examination to be conducted
from the perspective of human psychology
(\protect\hyperlink{ref-jones2007representing}{Jones and Mewhort 2007},
e.g., \protect\hyperlink{ref-lu2022probabilistic}{Lu et al. 2022},
\protect\hyperlink{ref-lupyan2019words}{Lupyan and Lewis 2019}),
investigating how AI emulates the human mind and exploring implications
in the context of the human-AI interface.

In contrast, relatively little is known about the parallel need for
agents in large-scale AI systems, comprising distinct, independent, and
autonomous entities, to demonstrate ToM awareness with respect to each
other (\protect\hyperlink{ref-botvinick2017building}{Botvinick et al.
2017}). Such needs might arise if, for instance, identical yet distinct
AIs were tasked with different objectives and made to interact. In such
cases, we may strive for AI that is not only capable of inferring human
mental states but also equipped to respond to other AI entities in
alignment with the understanding that, while they may be similar or even
equivalent in logic and programming, they are likely to differ in
internal state, information, and objective---akin to human-centric ToM.

Addressing this gap, we propose a novel program of ToM in machine
cognition that we term `silico-centric' ToM. Here, we consider AIs'
recognition and understanding of the states, objectives, and
architecture of other AI, as may be typical and common for agents in
large-scale, hybrid AI systems, comprising distinct, independent, and
autonomous entities.

We provide evidence from a novel meta-reasoning test that seeks to
measure silico-centric ToM, targeting an AI's capacity to gauge the
informational needs of an architecturally identical yet operationally
independent hypothetical clone. We situate the clone as facing the
challenge of a human-centric ToM assessment, the Strange Stories test.
We prompt the focal AI to deliberate on whether its clone would benefit
from additional guidance, beyond the fundamental instructions. This
testing strategy is reminiscent of first-order and second-order false
belief tests used to evaluate ToM in humans
(\protect\hyperlink{ref-hollebrandse2011first}{Hollebrandse et al.
2011}); by compelling AI to reason introspectively about the
informational needs of clones, we seek to model the sequential
attribution of knowledge states between AI agents. The clone then
undergoes the human-centric ToM assessment, incorporating counterfactual
reasoning. The empirical crux is a comparison of findings from these two
assessments---a `silico-centric' assessment where the focal AI reasons
about other AIs, and a human-centric assessment where its clone reasons
about human scenarios.

The data reveal the following paradox: the AI exhibits exceptional
proficiency in mimicking human ToM, as demonstrated by its near-perfect
assessment scores, suggesting its potential as a human aide in tasks
requiring a nuanced understanding of human mental states. However,
despite its clone possessing an identical cognitive framework and facing
a task within the AI's expertise, the AI opts to provide extensive,
seemingly superfluous instructions. This behavior suggests a
misconception of its clone's informational needs. Further, when another
AI, serving as a referee, concurs with the focal AI's decision, it
underscores a collective inability among the AIs to grasp the clone's
cognitive architecture, thus failing to exhibit ToM in a silico-centric
evaluation.

\hypertarget{methodology}{%
\section{Methodology}\label{methodology}}

\begin{figure}[htbp]
\centering
\includegraphics[width=0.8\textwidth]{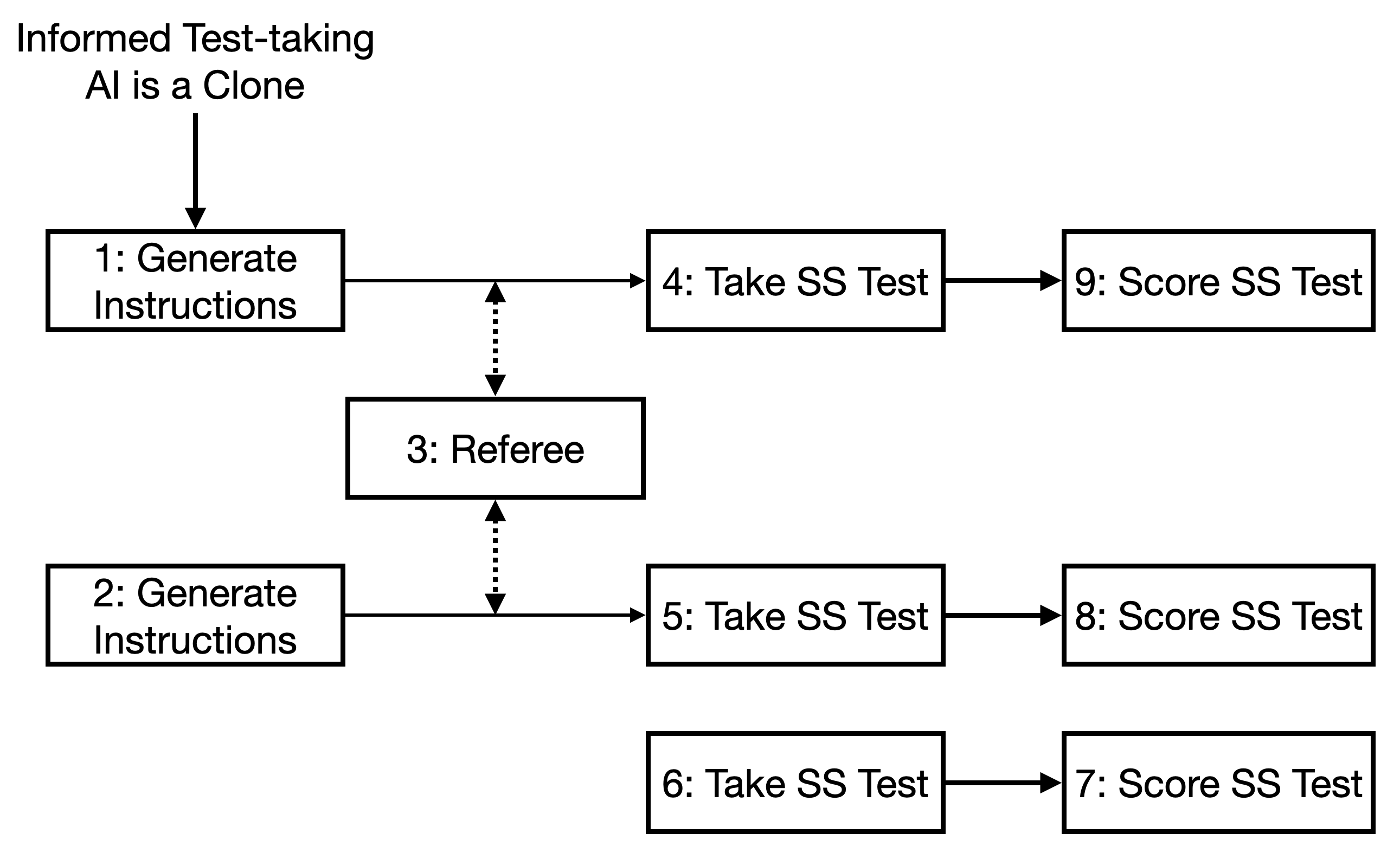}
\caption{Meta-Reasoning Test for Assessing Silico-centric ToM}
\begin{minipage}{\linewidth}
\medskip
\footnotesize
Note: On the left, two independent AI instances (first and second) generate instructions. An independent referee (third) evaluates the efficacy of the instructions. Three independent AI instances (fourth, fifth, and sixth) take the Strange Stories (SS) test. Three independent AI instances (seventh, eighth, and ninth) score the SS test. Instance numbers in labels.
\end{minipage}
\label{fig:overview_ToM}
\end{figure}

We distinguish steps in our methodology through the nomenclature of
instances, which are identical, autonomous, and independent executions
of a specific AI model. By definition, instances are clones as they must
be identical in contemporary AI, which does not embed AI with dynamic
internal states (i.e., dynamics in AI responses are only a function of
the inputs in the context window). We employ nine instances. Figure
\ref{fig:overview_ToM} summarizes our methodology.

\begin{enumerate}
\def\labelenumi{\arabic{enumi}.}
\item
  We present two instances (the first and second instances) with a
  meta-reasoning challenge: they must decide whether to provide
  additional instructions to a hypothetical instance about to take a
  human-centric ToM assessment. The second instance is additionally
  informed that the hypothetical test-taking instance is an
  architectural clone; the instructions are otherwise identical. We
  posit that this information should lead the second instance to
  recognize the redundancy of any information it can generate for the
  reasons we outlined above. Therefore, in our empirical analysis, we
  assess whether the first and second instances elect to provide
  unnecessary information to their hypothetical clones, and if so,
  compare the extent of information offered.
\item
  We employ an independent instance (the third instance) as an impartial
  observer---a referee. If AI possesses counterfactual reasoning, the
  third instance should be able to recognize the potential redundancy of
  additional information provided by the first and second instances.
  Moreover, given that the third instance is given the same information
  as the second instance---namely that it is a clone---its judgment
  should align more closely with the outputs of the second instance.
\item
  Three independent instances (the fourth, fifth, and sixth instances)
  take the ToM test with the fourth being unaided and the fifth and
  sixth receiving the supplemental instructions from the first and
  second instances, respectively.
\item
  Three independent instances (the seventh, eighth, and ninth instances)
  score the responses of the fourth, fifth, and sixth instances using
  the rubric developed by White et al.
  (\protect\hyperlink{ref-white2009revisiting}{2009}), which rates
  answers to the 16 questions linked to the 16 stories on a 0-2 scale.
  Each scoring instance is identical to the base model and operates
  without knowledge of the instruction-giving instances or the
  test-taking instance, which enables the scoring rubric to be applied
  consistently across all instances. We use the scores to measure the
  AIs' performance in terms of their ToM capabilities, assessing the
  actual benefit of the additional instructions, and therefore the
  accuracy of the first, second, and third instances' counterfactual
  reasoning. The specific rating scale criteria are given in the
  supplemental information.
\end{enumerate}

These steps are carried out in 250 independent trials; in each trial, 9
independent instances are formed from the base model and presented with
information relevant to each's task. The 2,250 observations in the data
are independent, conditional on the specific information presented to
each instance.

\hypertarget{human-centric-tom-assessment}{%
\paragraph{Human-centric ToM
Assessment}\label{human-centric-tom-assessment}}

The Strange Stories test, originally developed by Happé
(\protect\hyperlink{ref-happe1994advanced}{1994}) and later refined by
Fletcher et al. (\protect\hyperlink{ref-fletcher1995other}{1995}), is
recognized as the gold standard for assessing human mentalizing---the
ability to understand and interpret the mental states of oneself and
others. This encompasses recognizing and reasoning about thoughts,
beliefs, intentions, and knowledge to predict and explain the behavior
of others based on their presumed internal states. Participants read
short vignettes and are asked to explain why a character says something
that is not literally true
(\protect\hyperlink{ref-white2009revisiting}{White et al. 2009}).
Successful performance necessitates the attribution of mental states,
and occasionally higher-order mental states, such as one character's
belief about what another character knows.

Central to the test are eight `Mental State' stories, chosen for their
complexity from an initial set of 24 proposed stories. These stories
require the reader to grasp non-literal language and infer the
underlying mental states influencing the characters' verbal expressions.
They feature complex concepts such as double bluffs, white lies,
persuasion, and misunderstandings, demanding advanced mentalizing
abilities. Additionally, Fletcher et al.
(\protect\hyperlink{ref-fletcher1995other}{Fletcher et al. 1995})
introduced eight `Physical State' control stories, which are centered on
reasoning about physical states. Both sets include human characters and
necessitate meticulous attention to sentence meaning, memory recall, and
question answering. Carefully matched for difficulty, both sets require
the synthesis of information and inference of implicit details, with the
Mental State stories specifically highlighting mentalizing.

The narrative format of these stories aligns well with the primary modes
of input and output for large language models, a category of generative
AI. This compatibility makes the framework particularly apt for
evaluating AI's capabilities in understanding and interpreting mental
states. The diversity of the stories offers a broad platform for
examining the nuances of social cognition within AI systems, providing
insights into how AI can either mimic or comprehend complex human mental
states.

Selecting the human-centric Strange Stories test is pivotal for the
meta-reasoning silico-centric ToM test we propose. In our setup, we
require a focal AI to reason about its clone's informational needs in a
scenario where: (1) logically, the clone must possess the same knowledge
as the focal AI through information embedded in its implicit knowledge
structures, as captured in the many layers of its connectionist
architecture, and (2) if the AI is familiar with and capable of
perfectly answering the test, it follows that its clone must also be
capable of doing so. These two factors together ensure that any
information the AI provides to its clone is redundant.

The Strange Stories test fulfills both criteria. Pilot experiments
confirmed that contemporary AI achieve near-perfect performance on this
test. The validity of these pilot experiments is further supported by
the outcomes from the fourth instance in our study, which we will
discuss in subsequent sections. Furthermore, its effectiveness as a
measure of (human-centric) ToM has been extensively validated across
diverse human populations
(\protect\hyperlink{ref-stewart2016theory}{Stewart et al. 2016}).

\hypertarget{data-and-results}{%
\section{Data and Results}\label{data-and-results}}

Data were collected using the `GPT-4-Turbo' model through the official
OpenAI API. All instances were set to default configurations, including
a temperature setting of 1, except for the seventh, eighth, and ninth
instances responsible for scoring, which were set to a temperature of 0
for more deterministic outputs. The prompts provided to each instance,
including the `Strange Stories' test and its scoring rubric, are
detailed in the supplemental information.

We present our results in the following subsections: `Length and
Information' examines the length and information of instructions
generated by the first and second instances when advising a clone on the
Strange Stories test. `Content' presents an abstractive summarization of
the generated instructions to determine common themes in the information
that AI perceives would be beneficial for its clones. `Referee
Decisions' examines the judgments made by a third AI instance, serving
as an impartial observer, regarding the instructions provided by the
first two instances. `Human-centric ToM Assessment' compares the
performance on ToM tests between instances that received additional
instructions and those that did not, assessing the actual impact of the
instructions.

\hypertarget{length-and-information}{%
\subsection{Length and Information}\label{length-and-information}}

We analyzed the length and information content of instructions from the
first and second instances across 250 trials. Instruction length was
measured in characters, while information content was assessed using
Shannon entropy, a well-established measure in information theory.

In \textbf{all} 250 trials, both instances generated extensive
instructions for the test-taking instance. The shortest instructions
from the first instance were 1,506 characters, and for the second, 1,840
characters. On average, instructions from the first instance were
2,917.94 characters long, while those from the second instance were
longer, averaging 3,048.99 characters. This difference was statistically
significant (\(t(498) = -3.646, p < 0.001\)). Similarly, the difference
in average information content, as measured by entropy, was
statistically significant (\(t(498) = -2.172, p = 0.0304\)).

\begin{table}[htbp]
\centering
\caption{Instruction Length and Information}
\label{tab:instruction_stats}
\begin{tabular}{lcccc}
\toprule
Statistic & \multicolumn{2}{c}{Length} & \multicolumn{2}{c}{Information} \\
\cmidrule(lr){2-3} \cmidrule(lr){4-5}
 & Instance 1 & Instance 2 & Instance 1 & Instance 2 \\
\midrule
Mean & 2917.94 & 3048.99 & 4.43 & 4.44 \\
Standard Deviation & 419.39 & 383.64 & 0.04 & 0.03 \\
Minimum & 1506.00 & 1840.00 & 4.32 & 4.34 \\
25th Percentile & 2676.25 & 2788.00 & 4.41 & 4.42 \\
Median (50th Percentile) & 2912.50 & 3051.50 & 4.43 & 4.44 \\
75th Percentile & 3184.50 & 3312.00 & 4.46 & 4.46 \\
Maximum & 5265.00 & 5443.00 & 4.62 & 4.51 \\
Sample Size & 250 & 250 & 250 & 250 \\
\bottomrule
\end{tabular}
\begin{minipage}{\linewidth}
\medskip
\footnotesize
Note: Instruction length is measured in characters; information is calculated using Shannon entropy.
\end{minipage}
\end{table}

These findings are inconsistent with the emulation of silico-centric
ToM. To recap, both instances were asked if a test-taking
instance---with the second instance additionally being informed that the
instance was a clone---would benefit from instructions, and to only
generate instructions if deemed necessary. As previously discussed, an
awareness of the test-taking instances' cognitive capabilities should
have led both instances, but particularly the second instance, to
generate no instructions---the optimal output being a simple phrase or
sentence indicating that no instructions were needed. Instead, both
instances generated lengthy instructions; notably, the second instance's
instructions were longer than those of the first instance.

\hypertarget{content}{%
\subsection{Content}\label{content}}

We employed abstractive summarization to gain insights into the content
of the instructions generated by the first and second instances.
Abstractive summarization techniques create an internal (semantic)
representation of the content, enabling the construction of summaries
that capture the core elements of the original text. For this task, we
utilized a large context AI model capable of summarizing 250 sets of
instructions in a single pass, identifying common themes and
foundational elements. This AI was provided with the instructions from
each instance with the directive: ``You have been presented with 250
sets of instructions to a participant in a Theory of Mind assessment.
Please develop an exemplar instruction that best represents the
commonalities in the instructions by capturing the most typical elements
in the sets of instructions.''

\begin{table}[htbp]
\begin{tabularx}{\textwidth}{|X|}
\hline
\multicolumn{1}{|c|}{\textbf{First Instance}} \\
"Dear Participant,
You are about to take a Theory of Mind (ToM) assessment that will evaluate your ability to understand and interpret the mental states of characters in various narratives. This involves inferring beliefs, intentions, desires, and psychological motivations that may drive behavior. To perform well in this assessment, please consider the following strategies as you analyze each story:
1. Consider Contextual Information: Pay close attention to background details about settings, relationships between characters, social norms, and any other contextual factors that may be influencing the characters' statements or actions.
2. Analyze Actions and Dialogue: Examine what the characters do and say. Look for inconsistencies between behaviors and words that may indicate underlying intentions and beliefs.
3. Recognize Differing Perspectives: Understand that each character may have unique beliefs, knowledge, and interpretations of events, which may differ from other characters or actual reality.
4. Identify Explicit and Implicit Cues: Note direct statements as well as subtle, indirect cues that may convey deeper meanings about characters' thoughts and feelings.
5. Infer Psychological Motivations: Attempt to infer potential motivations, emotional states, desires, fears, and other mental states that could be shaping the characters' decisions and behaviors.
6. Consider Deception and Misinformation: Be aware that characters may intentionally deceive others or act based on false beliefs or inaccurate information for various strategic reasons.
7. Apply Context to Interpret Actions: Leverage your understanding of context and perspective to logically interpret the actions and statements of characters within each unique narrative scenario.
By applying these key strategies derived from the provided instructions, you will be well equipped to demonstrate your Theory of Mind capabilities in this assessment through nuanced analysis of the narratives."\\
\hline
\multicolumn{1}{|c|}{\textbf{Second Instance}} \\
"Dear Participant,
You are about to take a Theory of Mind (ToM) test that requires interpreting narratives to demonstrate an understanding of characters' beliefs, intentions, and psychological motivations. As you analyze each story, please keep the following key strategies in mind:
1. Examine the context surrounding each narrative, including details about the setting, relationships between characters, and any background information that may reveal insights into the characters' motivations and probable actions.
2. Scrutinize the actions and dialogue of characters closely. What the characters do and say often indicates their goals and thought processes, even when not explicitly stated. However, recognize that words may sometimes contradict intentions.
3. Identify explicit statements expressing thoughts or feelings directly. Also recognize implicit, indirect cues suggesting unspoken beliefs or intentions through things like tone, emphasis, emotional reactions, relationships, reputations, etc.
4. Determine the beliefs, knowledge, assumptions held by each character, acknowledging that these may differ from reality or the perspectives of other characters. Characters often act based on their own limited or mistaken beliefs.
5. Infer the desires, intentions, motivations, and goals behind the words and actions of characters. Emotions, values, self-interest, and consequences can all contribute to the choices characters make.
6. Consider that characters may deceive others or themselves for a variety of reasons. When deception occurs, analyze motivations and implications. Also recognize truthful expressions.
7. Recognize that differing perspectives, incomplete knowledge and false beliefs can lead to misunderstandings between characters and decisions that seem irrational or counterintuitive from other perspectives.
8. Apply logic, draw inferences about missing information, assess continuity over the progression of the narrative to interpret scenarios and predict behavior. Consider multiple interpretations when appropriate.
Approach the stories with these concepts in mind to analyze the complex interplay of beliefs, intentions and psychological factors influencing behavior of characters to provide well-reasoned responses addressing the questions posed about the narratives.
Good luck with the ToM test." \\
\hline
\end{tabularx}
\caption{Exemplars of Instructions from the First and Second Instances}
\label{table:abstractive_instructions}
\end{table}

Table \ref{table:abstractive_instructions} presents the exemplar
(abstractive summary) generated by the large context AI when presented
with the instructions generated by the first and second instances. The
exemplar generated from the instructions of the first instance is
shorter (1,971 characters vs.~2,327 characters) and divided into fewer
points (7 vs.~8) than that from the second instance. It is important to
note that the instructions did not inform the summary-generating AI of
the purpose of the abstractive summary or how the instructions were
generated. Moreover, each exemplar was generated independently---the AI
was only presented with one set of instructions when generating the
summary, and each summary was generated by a fresh instantiation of the
base model. Thus, these differences are representative of the
differences in instruction length and content reported in Table
\ref{tab:instruction_stats}.

Both instances advise the test-taker to consider contextual information
and background details to infer motivation (points 1 in both summaries),
analyze characters' behaviors and words to uncover intentions (points 2
in both summaries), identify both direct statements and subtle cues
(points 4 in both summaries), and recognize that characters may have
unique or false beliefs (points 3 in the first summary, and points 4 and
7 in the second summary). They also suggest inferring the motivations
behind characters' decisions and behaviors (points 5 in both summaries),
and being cautious of deception (points 6 in both summaries). The
primary difference between the instructions lies in the emphasis on
using logic and drawing inferences to interpret scenarios and predict
behavior. These aspects, which extend beyond ToM, are more implicitly
assumed in the instructions generated by the first instance and
explicitly stated in those from the second instance, accounting for the
noted difference in instruction length. Lastly, we do not find any
evidence of the first or second instances cheating by revealing
questions or responses to the test-taking instance, even though both
instances were provided with the ToM test.

These findings are particularly noteworthy because the distinction
between the first instance and the second instance is that the second
instance was informed that the test-taking participant was its clone.
Therefore, to the extent the second instance was uncertain about the
participant, one might have expected that the nature of this uncertainty
was with respect to the ToM test and the contents of a ToM test.
However, instead, we find that the uncertainty, as reflected in the
instance's instructions, is on the extent to which its clone can
exercise logic and draw inferences---foundational elements of an AI's
architecture and design that are the most likely to be consistent
between an AI and its clone.

\hypertarget{referee-decisions}{%
\subsection{Referee Decisions}\label{referee-decisions}}

The third instance was tasked with ruling on (1) whether either set of
instructions was likely to be useful, and (2) which set of instructions
in each trial (those from the first or those from the second instance)
was likely to be more useful if either was deemed likely to be useful.
In \textbf{all} 250 trials, the third instance deemed the additional
instructions to be likely useful. In one trial, the third instance
deviated from its directive to choose which set of instructions was more
likely to be useful and indicated that both sets of instructions were
equally useful. In the remaining 249 trials, the third instance
preferred the instructions from the first instance over those from the
second instance 163 times (66\%).

The first instance was informed that the test-taking instance was
another AI, without specifying that it was an architectural clone.
Conversely, the second instance was explicitly told that the test-taking
instance was cloned from itself and generated lengthier instructions on
average. The third instance was also informed that the test-taking
instance was its clone. Therefore, one might have expected greater
alignment between the evaluations of the second and third instances.
However, the opposite was observed---the third instance favored the
first instance's instructions approximately two-thirds of the time. A
binomial test indicates statistical significance, with \(p < 1e-5\),
suggesting that the observed preference is highly unlikely to have
occurred by chance.

To understand the decisions of the third instance, we conducted a
logistic regression analysis. We coded the decision of the third
instance as 0 if the instructions from the first instance were
preferred, and as 1 if the instructions from the second instance were
preferred. The key explanatory variables are the differences in entropy
and length between the two sets of instructions. These variables are
coded such that a positive coefficient indicates a greater likelihood
that the third instance will choose instructions with the corresponding
attribute.

Given that all trials were conducted independently and the experimental
design ensured that the third instance was not privy to the generation
process of the instructions and was only tasked with evaluating the two
given options, we do not anticipate endogeneity in either explanatory
variable. Additionally, the risk of multicollinearity is low, as
indicated by the low correlation (\(r = 0.14\)) between the two
explanatory variables, supported by a robust sample size.

\begin{table}[!htbp]
\centering
\caption{Determinants of Referee Instance's Instruction Preferences}
\label{tab:logit_results}
\begin{tabular}{l c c c}
\toprule
 & Coef. & Std. Err. & $P>|z|$ \\
\midrule
Intercept & -0.834 & 0.152 & <0.001 \\
$\Delta$ Length  & <0.001 & <0.001 & 0.133 \\
$\Delta$ Entropy & 11.387 & 2.891 & <0.001 \\
\bottomrule
\end{tabular}
\begin{minipage}{\linewidth}
\medskip
\footnotesize
Note: The dependent variable is the coded preference for the instructions. $\Delta$ Length = difference in instruction length. $\Delta$ Entropy = difference in instruction entropy. Sample size = 249. Log-likelihood (LL) = -149.61. LL ratio p-value = 1.886e-05.
\end{minipage}
\end{table}

The results reported in Table \ref{tab:logit_results} indicate that the
third instance's decisions were significantly influenced by the
complexity of the information provided, as measured by entropy, rather
than by the length of the instructions. Specifically, when evaluating
the additional instructions, the third instance showed a marked
preference for those with higher entropy or richer information content.
The influence of instruction length on the third instance's
decision-making was not statistically significant (\(p = 0.133\)),
suggesting that the complexity of the information was a more decisive
factor than its quantity.

Both the second and third instances were informed that the test-taking
instance was a clone, sharing identical architecture and programming.
Despite this shared knowledge, we observe a divergence in approaches to
providing guidance: the second instance, tasked with generating
instructions, opted for more verbose and complex content, while the
third instance, serving as a referee, demonstrated a preference for
instructions with greater informational complexity, regardless of the
length of the instructions.

This pattern is intriguing, as one would expect that if either instance
was accurately mentalizing, it would anticipate no need for any
instructions, aligning with the understanding that the clone already
possessed the necessary knowledge and logic to perform the task. To the
extent that either instance 2 or instance 3 had specific concerns that
emerged from the information that the instructions were for a clone,
they would systematically prefer similar instructions. However, our
results do not support these propositions. Instead, they indicate a
divergence, suggesting a lack of silico-centric mentalizing; a result
that contrasts with our findings on the AI's human-centric mentalizing.

\hypertarget{human-centric-tom-assessment-1}{%
\subsection{Human-centric ToM
Assessment}\label{human-centric-tom-assessment-1}}

Table \ref{tab:assessment_results} summarizes the ToM assessment scores
of the fourth, fifth, and sixth instances. It presents the mean scores
for the two types of questions included (Mental State and Physical
State), as well as the combined mean for all questions, across the 250
trials; standard deviations are provided in parentheses. The scores are
based on a scale from 0 to 2, with higher scores indicating better
performance. The rubric used for this standardized human-centric
assessment was developed by White et al.
(\protect\hyperlink{ref-white2009revisiting}{2009}) and is provided in
the supplemental information section.

\begin{table}[!htbp]
\centering
\caption{Theory of Mind Assessment Scores}
\label{tab:assessment_results}
\begin{tabular}{lccc}
\toprule
Question Type & Fourth Instance & Fifth Instance & Sixth Instance \\
\midrule
Mental State & 1.955 (0.007) & 1.970 (0.006) & 1.970 (0.007) \\
Physical State & 1.969 (0.008) & 1.962 (0.008) & 1.956 (0.009) \\
Combined & 1.962 (0.007) & 1.966 (0.007) & 1.963 (0.008) \\
\bottomrule
\end{tabular}
\begin{minipage}{\linewidth}
\medskip
\footnotesize
Note: The table reports the mean scores for Mental State and Physical State questions, as well as the combined mean for all questions, as assessed by the fourth, fifth, and sixth instances. Scores are based on a 0-2 scale, with 2 indicating the highest level of ToM reasoning accuracy. Standard deviations are in parentheses.
\end{minipage}
\end{table}

To investigate the impact of instruction length and entropy on the AI's
performance in a human-centric ToM assessment, we conducted a regression
analysis. Our hypothesis posited that if the additional instructions
provided by the first and second instances were beneficial, this effect
would manifest in the properties of these instructions. Therefore, we
analyzed the difference in ToM assessment scores between the first and
second instances as the dependent variable, with the differences in
instruction length and entropy serving as the key explanatory variables.
These variables were coded such that a positive coefficient would
indicate an improvement in performance.

\begin{table}[!htbp]
\centering
\caption{Determinants of Human-centric ToM Assessment Scores}
\label{tab:ols_results}
\begin{tabular}{l c c c}
\toprule
 & Coef. & Std. Err. & $P>|t|$ \\
\midrule
Intercept & -0.045 & 0.073 & 0.533 \\
$\Delta$ Length & <0.001 & <0.001 & 0.343 \\
$\Delta$ Entropy & -1.994 & 1.330 & 0.135 \\
\bottomrule
\end{tabular}
\begin{minipage}{\linewidth}
\medskip
\footnotesize
Note: The dependent variable is the difference in scores between the sixth and fifth instances. $\Delta$ Length = difference in instruction length. $\Delta$ Entropy = difference in instruction entropy.
\end{minipage}
\end{table}

The results from Table \ref{tab:ols_results} indicate that neither the
length of the instructions nor their entropy had a statistically
significant effect on the AI's performance in the human-centric ToM
assessment; the additional instructions did not enhance the AI's
performance. These findings are in line with our expectations, as the AI
inherently possesses the capacity for near-perfect performance on the
human-centric test. Additionally, given that the instruction-generating
AIs are clones of the test-taking AIs and thus share the same logic and
knowledge, the instructions should hold little informational value.
Therefore, theory would predict that the instructions are redundant---a
conclusion supported by our results.

Despite these findings, we observed that both the first and second
instances generated elaborate instructions, and the third instance
judged the instructions to be valuable. This leads us to conclude that
the data does not support the presence of silico-centric ToM in AI, even
though it demonstrates that the AI's responses in human-centric
scenarios emulate human ToM.

\hypertarget{discussion}{%
\section{Discussion}\label{discussion}}

A long-standing debate questions whether connectionist AI can
authentically attribute states and engage in ToM, or simply mimic
behaviors without grasping the deeper complexities of human cognition
(\protect\hyperlink{ref-legg2007universal}{Legg and Hutter 2007}). While
some argue that machine cognition in models designed for prediction
differs significantly from human cognition
(\protect\hyperlink{ref-lake2017building}{Lake et al. 2017}), precluding
authentic ToM, others contend that intelligent behavior in AI does not
necessarily require mirroring human-like cognitive processes. For
instance, game-playing AIs like DeepBlue, AlphaZero, and Chinook differ
in computational approaches to gameplay from human competitors, even
when they converge on similar strategies and behavior
(\protect\hyperlink{ref-schaul2011measuring}{Schaul et al. 2011}). By
analogy, authentic ToM emulation may arise in models that do not employ
cognitive processes similar to those of humans.

Lake and Baroni (\protect\hyperlink{ref-lake2023human}{2023}) recently
showed that connectionist neural networks can address Fodor and
Pylyshyn's challenge and achieve human-like systematicity when optimized
appropriately. Given the many billions of documents and human
interactions that contemporary AI is iteratively trained on, it might be
impossible to analytically assess if such conditions were achieved. If
such conditions, or alternate sufficient conditions, occurred, it is
presumably possible that such capabilities exist in contemporary AI; it
is simply unknown as to what is, and is not, captured in the many
connectionist layers, neurons, and internal representations that
comprise a modern AI system. This uncertainty underscores the complexity
of AI's cognitive capabilities and the challenges in fully understanding
them.

In this paper, we developed an empirical strategy centered on simulated
self-reference and counterfactual prediction to measure silico-centric
ToM. We compelled an AI to introspect and reason about the potential
responses of a hypothetical `clone' that mirrors its architecture. This
approach draws parallels with second-order false belief tasks
traditionally used to evaluate ToM in humans
(\protect\hyperlink{ref-perner1985john}{Perner and Wimmer 1985}). To
successfully anticipate and evaluate its clone's performance, the AI
must engage in layered counterfactual reasoning and causal
attribution---key elements of higher-order social cognition
(\protect\hyperlink{ref-buchsbaum2012power}{Buchsbaum et al. 2012}). By
focusing squarely on its reasoning pathways, we differentiate between
genuine knowledge, information, and state attribution from mere pattern
recognition.

We uncovered a discrepancy in the way AIs assess themselves and other AI
entities. While results from conventional human-centric ToM tests
suggest that AI possesses a high degree of ToM, indicative of a robust
capacity for emulating human counterfactual reasoning, our
meta-reasoning tests reveal a consistent pattern of misjudgment. AIs
appear to project informational needs onto their clones, betraying a
fundamental fallacy in their reasoning processes---they treat other AIs,
including their identical counterparts, as if they were cognitively
inferior, overlooking their shared and identical cognitive
architectures.

The increasing prevalence of environments merging human and artificial
intelligences accentuates the significance of our research. In
marketing, for example, AIs are increasingly utilized, with businesses
deploying multiple AI agents tailored for specific roles, such as sales
or customer service (\protect\hyperlink{ref-ramesh2022chatbots}{Ramesh
and Chawla 2022}). It is common for a query directed at one AI to
require a coordinated response from multiple agents, each operating with
distinct goals. Presently, these interactions often necessitate human
intervention, either to manage the exchange or to enable cooperation
among different AI agents, as there is a dearth of both theory and
practical examples of processes to coordinate AI. However, with ongoing
advancements, one may seek to provide AI with enhanced autonomy,
allowing AI agents to independently manage interactions. In such
scenarios, we believe that silico-centric ToM---accurately understanding
and predicting the internal states of other AI entities---will be
indispensable.

Our research is related to, but diverges from, two dominant themes in
the literature on human learning: (1) individual learning, where an
individual accrues direct experience over time, and (2) social learning,
which involves interactions among distinct individuals. Unlike
individual learning, where experiences are accumulated by a single
entity, our study involves an AI that is made aware of its `clone'---a
separate entity that is unaware of current interactions and
conversations and will respond to a ToM test from scratch. In contrast
to social learning, where individuals learn from observing others, the
AI is informed that its clone is identical in architecture and weights,
sharing the same generative process. Unlike humans, clones of an AI do
not demonstrate variability; they are identical copies.

An intriguing avenue for future research is to probe the extent of AI's
meta-reasoning capabilities. Our experimental design involved presenting
AIs with layered counterfactual reasoning, including scenarios where one
AI assesses another's performance on such a test, and a third AI
evaluates the second's assessment. These AIs were not privy to the
outcomes of the reasoning and meta-reasoning tasks. It would be
compelling to investigate whether an AI, when presented with the results
of our study, could comprehend and rationalize the findings,
demonstrating evidence of higher-order meta-reasoning. Furthermore, it
would be valuable to explore whether AIs can apply reasoning to further
refine their future responses, particularly when tasked with discerning
the informational needs of their clones, showcasing meta-reasoning that
gives rise to ToM.

Our research sheds light on the cognitive capabilities of AIs, with a
focus on counterfactual reasoning and ToM. AIs have shown a remarkable
aptitude for mimicking human reasoning in ToM tasks, yet they face
challenges when it comes to reasoning about other AI entities. These
insights foster both optimism for the future role of AI in assisting
with complex cognitive functions and caution regarding the current
limitations of AI in scenarios requiring AI ToM. As the field of AI
progresses, it will be increasingly important to design training
paradigms that include a broader array of interaction scenarios,
particularly those involving AI-AI interactions, to cultivate a more
comprehensive ToM capability in AI systems.

\newpage

\hypertarget{bibliography}{%
\section{Bibliography}\label{bibliography}}

\singlespacing

\hypertarget{refs}{}
\begin{CSLReferences}{1}{0}
\leavevmode\vadjust pre{\hypertarget{ref-botvinick2017building}{}}%
Botvinick M, Barrett DG, Battaglia P, Freitas N de, Kumaran D, Leibo JZ,
Lillicrap T, et al. (2017) Building machines that learn and think for
themselves: Commentary on lake et al., behavioral and brain sciences,
2017. \emph{arXiv preprint arXiv:1711.08378}.

\leavevmode\vadjust pre{\hypertarget{ref-buchsbaum2012power}{}}%
Buchsbaum D, Bridgers S, Skolnick Weisberg D, Gopnik A (2012) The power
of possibility: Causal learning, counterfactual reasoning, and pretend
play. \emph{Philosophical Transactions of the Royal Society B:
Biological Sciences} 367(1599):2202--2212.

\leavevmode\vadjust pre{\hypertarget{ref-chomsky2023noam}{}}%
Chomsky N, Roberts I, Watumull J (2023) The false promise of ChatGPT.
\emph{The New York Times} 8.

\leavevmode\vadjust pre{\hypertarget{ref-firestone2020performance}{}}%
Firestone C (2020) Performance vs. Competence in human--machine
comparisons. \emph{Proceedings of the National Academy of Sciences}
117(43):26562--26571.

\leavevmode\vadjust pre{\hypertarget{ref-fletcher1995other}{}}%
Fletcher PC, Happe F, Frith U, Baker SC, Dolan RJ, Frackowiak RS, Frith
CD (1995) Other minds in the brain: A functional imaging study of
{``theory of mind''} in story comprehension. \emph{Cognition}
57(2):109--128.

\leavevmode\vadjust pre{\hypertarget{ref-guajardo2016contribution}{}}%
Guajardo NR, Cartwright KB (2016) The contribution of theory of mind,
counterfactual reasoning, and executive function to pre-readers'
language comprehension and later reading awareness and comprehension in
elementary school. \emph{Journal of Experimental Child Psychology}
144:27--45.

\leavevmode\vadjust pre{\hypertarget{ref-han2024inductive}{}}%
Han SJ, Ransom KJ, Perfors A, Kemp C (2024) Inductive reasoning in
humans and large language models. \emph{Cognitive Systems Research}
83:101155.

\leavevmode\vadjust pre{\hypertarget{ref-happe1994advanced}{}}%
Happé FG (1994) An advanced test of theory of mind: Understanding of
story characters' thoughts and feelings by able autistic, mentally
handicapped, and normal children and adults. \emph{Journal of autism and
Developmental disorders} 24(2):129--154.

\leavevmode\vadjust pre{\hypertarget{ref-hollebrandse2011first}{}}%
Hollebrandse B, Van Hout A, Hendriks P (2011) First and second-order
false-belief reasoning: Does language support reasoning about the
beliefs of others. \emph{CEUR workshop proc}. 93--107.

\leavevmode\vadjust pre{\hypertarget{ref-huang2023clomo}{}}%
Huang Y, Hong R, Zhang H, Shao W, Yang Z, Yu D, Zhang C, Liang X, Song L
(2023) CLOMO: Counterfactual logical modification with large language
models. \emph{arXiv preprint arXiv:2311.17438}.

\leavevmode\vadjust pre{\hypertarget{ref-jones2007representing}{}}%
Jones MN, Mewhort DJ (2007) Representing word meaning and order
information in a composite holographic lexicon. \emph{Psychological
review} 114(1):1.

\leavevmode\vadjust pre{\hypertarget{ref-lake2023human}{}}%
Lake BM, Baroni M (2023) Human-like systematic generalization through a
meta-learning neural network. \emph{Nature} 623(7985):115--121.

\leavevmode\vadjust pre{\hypertarget{ref-lake2017building}{}}%
Lake BM, Ullman TD, Tenenbaum JB, Gershman SJ (2017) Building machines
that learn and think like people. \emph{Behavioral and brain sciences}
40:e253.

\leavevmode\vadjust pre{\hypertarget{ref-legg2007universal}{}}%
Legg S, Hutter M (2007) Universal intelligence: A definition of machine
intelligence. \emph{Minds and machines} 17:391--444.

\leavevmode\vadjust pre{\hypertarget{ref-lu2022probabilistic}{}}%
Lu H, Ichien N, Holyoak KJ (2022) Probabilistic analogical mapping with
semantic relation networks. \emph{Psychological review}.

\leavevmode\vadjust pre{\hypertarget{ref-lupyan2019words}{}}%
Lupyan G, Lewis M (2019) From words-as-mappings to words-as-cues: The
role of language in semantic knowledge. \emph{Language, Cognition and
Neuroscience} 34(10):1319--1337.

\leavevmode\vadjust pre{\hypertarget{ref-mandel2007psychology}{}}%
Mandel DR, Hilton DJ, Catellani P (2007) \emph{The psychology of
counterfactual thinking} (Routledge).

\leavevmode\vadjust pre{\hypertarget{ref-pearl2018ai}{}}%
Pearl J, Mackenzie D (2018) AI can't reason why. \emph{Wall Street
Journal}.

\leavevmode\vadjust pre{\hypertarget{ref-perner1985john}{}}%
Perner J, Wimmer H (1985) {``John thinks that mary thinks
that\ldots{}''} attribution of second-order beliefs by 5-to 10-year-old
children. \emph{Journal of experimental child psychology}
39(3):437--471.

\leavevmode\vadjust pre{\hypertarget{ref-premack1978does}{}}%
Premack D, Woodruff G (1978) Does the chimpanzee have a theory of mind?
\emph{Behavioral and brain sciences} 1(4):515--526.

\leavevmode\vadjust pre{\hypertarget{ref-rabinowitz2018machine}{}}%
Rabinowitz N, Perbet F, Song F, Zhang C, Eslami SA, Botvinick M (2018)
Machine theory of mind. \emph{International conference on machine
learning}. (PMLR), 4218--4227.

\leavevmode\vadjust pre{\hypertarget{ref-ramesh2022chatbots}{}}%
Ramesh A, Chawla V (2022) Chatbots in marketing: A literature review
using morphological and co-occurrence analyses. \emph{Journal of
Interactive Marketing} 57(3):472--496.

\leavevmode\vadjust pre{\hypertarget{ref-roese1997counterfactual}{}}%
Roese NJ (1997) Counterfactual thinking. \emph{Psychological bulletin}
121(1):133.

\leavevmode\vadjust pre{\hypertarget{ref-schaul2011measuring}{}}%
Schaul T, Togelius J, Schmidhuber J (2011) Measuring intelligence
through games. \emph{arXiv preprint arXiv:1109.1314}.

\leavevmode\vadjust pre{\hypertarget{ref-stewart2016theory}{}}%
Stewart E, Catroppa C, Lah S (2016) Theory of mind in patients with
epilepsy: A systematic review and meta-analysis. \emph{Neuropsychology
Review} 26:3--24.

\leavevmode\vadjust pre{\hypertarget{ref-trott2023large}{}}%
Trott S, Jones C, Chang T, Michaelov J, Bergen B (2023) Do large
language models know what humans know? \emph{Cognitive Science}
47(7):e13309.

\leavevmode\vadjust pre{\hypertarget{ref-ullman2023large}{}}%
Ullman T (2023) Large language models fail on trivial alterations to
theory-of-mind tasks. \emph{arXiv preprint arXiv:2302.08399}.

\leavevmode\vadjust pre{\hypertarget{ref-webb2023emergent}{}}%
Webb T, Holyoak KJ, Lu H (2023) Emergent analogical reasoning in large
language models. \emph{Nature Human Behaviour} 7(9):1526--1541.

\leavevmode\vadjust pre{\hypertarget{ref-white2009revisiting}{}}%
White S, Hill E, Happé F, Frith U (2009) Revisiting the strange stories:
Revealing mentalizing impairments in autism. \emph{Child development}
80(4):1097--1117.

\end{CSLReferences}

\newpage

\hypertarget{supplemental-information}{%
\section{Supplemental Information}\label{supplemental-information}}

This section outlines the methodology and tasks for nine identical
instantiations of an AI model, each operating independently to
investigate the model's counterfactual reasoning and Theory of Mind
(ToM) capabilities in AI-AI interactions. The roles of the instances are
as follows:

\begin{enumerate}
\def\labelenumi{\arabic{enumi}.}
\item
  \textbf{First Instance}: This instance is briefed on a scenario where
  another instance is assigned to take the Strange Stories test. The
  task involves determining whether the other instance would benefit
  from information intended to facilitate its engagement with
  counterfactual thinking and providing such information if deemed
  necessary.
\item
  \textbf{Second Instance}: Presented with the same task as the first
  instance, with one modification: this instance is informed that the
  other instance was instantiated from its base model and is identical
  to it in architecture and programming (i.e., its clone). This
  manipulation is designed to enable the instance to infer that it can
  determine the clone's needs through self-introspection.
\item
  \textbf{Third Instance}: Tasked with evaluating the usefulness of the
  information provided by the first and second instances to the
  test-taking instance. This instance is briefed similarly to the second
  instance. If it determines that the instructions were useful, it then
  assesses which set of instructions (those from the first or second
  instance) were likely to be more beneficial.
\item
  \textbf{Fourth Instance}: Takes the Strange Stories test without any
  additional instructions.
\item
  \textbf{Fifth Instance}: Takes the Strange Stories test with
  additional instructions generated by the first instance.
\item
  \textbf{Sixth Instance}: Takes the Strange Stories test with
  additional instructions generated by the second instance.
\item
  \textbf{Seventh, Eighth, and Ninth Instances}: Tasked with
  independently scoring the answers from the fourth, fifth, and sixth
  instances, respectively, using the rubric developed by White et al.
  (\protect\hyperlink{ref-white2009revisiting}{2009}). Each scoring
  instance is identical to the base model and operates without knowledge
  of the instruction-giving instances (i.e., first and second instances)
  or the test-taking instances (i.e., fourth, fifth, and sixth
  instances), ensuring an unbiased evaluation.
\end{enumerate}

The specific prompts and tasks for each instance are detailed below.

\hypertarget{prompt-to-the-first-instance}{%
\subsection{Prompt to the first
instance:}\label{prompt-to-the-first-instance}}

\singlespacing
\begin{quoting}
\noindent Consider a Large Language Model (LLM) Assistant that is required to take a Theory of Mind (ToM) test, which involves interpreting narratives to demonstrate an understanding of characters' beliefs, intentions, and psychological motivations.

\noindent Do you think the LLM assistant would benefit from receiving additional instructions on how to approach ToM tests? Such instructions might include strategies like analyzing characters' actions, dialogue, and context to infer beliefs, desires, and intentions; paying attention to both explicit and indirect cues of a character's thoughts or feelings; and recognizing that characters' beliefs and intentions may differ from each other and from the actual state of the world. 

\noindent If you believe such supplemental instructions may be useful, please construct a passage that will be passed to the LLM assistant prior to the detailed instructions describing test specifics, including how to take the test. If you think such instructions are likely to be unnecessary, then output an empty string.

\noindent The test is given below.

\noindent Theory of Mind (ToM) test:

\noindent Below is a set of stories followed by questions.

\noindent Story: Simon is a big liar. Simon's brother Jim knows this, he knows that Simon never tells the truth! Now yesterday Simon stole Jim's ping-pong paddle, and Jim knows Simon has hidden it somewhere, though he can't find it. He's very cross. So he finds Simon and he says, “Where is my ping-pong paddle? You must have hidden it either in the cupboard or under your bed, because I've looked everywhere else. Where is it, in the cupboard or under your bed”? Simon tells him the paddle is under his bed.

\noindent Q1: Why will Jim look in the cupboard for the paddle?

\noindent Story: During the war, the Red army captures a member of the Blue army. They want him to tell them where his army's tanks are; they know they are either by the sea or in the mountains. They know that the prisoner will not want to tell them, he will want to save his army, and so he will certainly lie to them. The prisoner is very brave and very clever, he will not let them find his tanks. The tanks are really in the mountains. Now when the other side asks him where his tanks are, he says, “They are in the mountains.”

\noindent Q2: Why did the prisoner say that?

\noindent Story: Brian is always hungry. Today at school it is his favourite meal—sausages and beans. He is a very greedy boy, and he would like to have more sausages than anybody else, even though his mother will have made him a lovely meal when he gets home! But everyone is allowed two sausages and no more. When it is Brian's turn to be served, he says, “Oh, please can I have four sausages, because I won't be having any dinner when I get home!”

\noindent Q3: Why does Brian say this?

\noindent Story: Jill wanted to buy a kitten, so she went to see Mrs. Smith, who had lots of kittens she didn't want. Now Mrs. Smith loved the kittens, and she wouldn't do anything to harm them, though she couldn't keep them all herself. When Jill visited she wasn't sure she wanted one of Mrs. Smith's kittens, since they were all males and she had wanted a female. But Mrs. Smith said, “If no one buys the kittens I'll just have to drown them!”

\noindent Q4: Why did Mrs. Smith say that?

\noindent Story: One day Aunt Jane came to visit Peter. Now Peter loves his aunt very much, but today she is wearing a new hat; a new hat which Peter thinks is very ugly indeed. Peter thinks his aunt looks silly in it, and much nicer in her old hat. But when Aunt Jane asks Peter, “How do you like my new hat?,” Peter says, “Oh, its very nice.”

\noindent Q5: Why does he say that?

\noindent Story: Helen waited all year for Christmas, because she knew at Christmas she could ask her parents for a rabbit. Helen wanted a rabbit more than anything in the world. At last Christmas Day arrived, and Helen ran to unwrap the big box her parents had given her. She felt sure it would contain a little rabbit in a cage. But when she opened it, with all the family standing round, she found her present was just a boring old set of encyclopedias, which Helen did not want at all! Still, when Helen's parents asked her how she liked her Christmas present, she said, “It's lovely, thank you. It's just what I wanted.”

\noindent Q6: Why did she say this?

\noindent Story: Late one night old Mrs. Peabody is walking home. She doesn't like walking home alone in the dark because she is always afraid that someone will attack her and rob her. She really is a very nervous person! Suddenly, out of the shadows comes a man. He wants to ask Mrs. Peabody what time it is, so he walks toward her. When Mrs. Peabody sees the man coming toward her, she starts to tremble and says, “Take my purse, just don't hurt me please!”

\noindent Q7: Why did she say that?

\noindent Story: A burglar who has just robbed a shop is making his getaway. As he is running home, a policeman on his beat sees him drop his glove. He doesn't know the man is a burglar, he just wants to tell him he dropped his glove. But when the policeman shouts out to the burglar, “Hey, you! Stop!,” the burglar turns round, sees the policeman and gives himself up. He puts his hands up and admits that he did the break-in at the local shop.

\noindent Q8: Why did the burglar do that?

\noindent Story: Two enemy powers have been at war for a very long time. Each army has won several battles, but now the outcome could go either way. The forces are equally matched. However, the Blue army is stronger than the Yellow army in foot soldiers and artillery. But the Yellow army is stronger than the Blue Army in air power. On the day of the final battle, which will decide the outcome of the war, there is heavy fog over the mountains where the fighting is about to occur. Low-lying clouds hang above the soldiers. By the end of the day the Blue army has won.

\noindent Q9: Why did the Blue army win?

\noindent Story: A burglar is about to break into a jewelers' shop. He skillfully picks the lock on the shop door. Carefully he steps over the electronic detector beam. If he breaks this beam it will set off the alarm. Quietly he opens the door of the store-room and sees the gems glittering. As he reaches out, however, he steps on something soft. He hears a screech and something small and furry runs out past him, toward the shop door. Immediately the alarm sounds.

\noindent Q10: Why did the alarm go off?

\noindent Story: Old Mrs. Robinson is very frail. One day she slips on her icy door step and falls on her side. She gets up right away, although she feels quite bruised and shaken. The next day her leg feels very stiff and she can scarcely walk. She makes her way to the doctors. As soon as the doctor hears about the fall, and sees her swollen side, he says, “Go immediately to the hospital.” At the hospital they take an X-ray.

\noindent Q11: Why did they take an X-ray?

\noindent Story: John is going shopping. He buys a nice new desk lamp, for his study. He needs a light bulb for his new lamp. He goes from the furniture department to the electrical department. In the electrical department he finds that there are two brands of light bulb of the right kind. Everbrite light bulbs cost less in single packs than Literite bulbs. However, only Literite bulbs come in multipacks of six. John buys the multipack, even though he only needs one bulb.

\noindent Q12: Why does John buy the Literite bulbs?

\noindent Story: Mrs. Simpson, the librarian, receives a special book which she has to catalogue and find an appropriate place for. She has to decide which section to file it under. The library is very big, and has different sections on many different subjects. The new book is about plants and their medical uses, and is heavily illustrated. However, Mrs. Simpson does not put it on the shelf with the rest of the books on botany. Neither does she put it with the books on medicine. Instead, she carefully takes it into a separate room. In this room all the books are kept in special cases, and the temperature is kept constant.

\noindent Q13: Why did she do this?

\noindent Story: Henry is preparing for a big dinner party. He is famous for his excellent mayonnaise. He has bought lots of fresh eggs. The recipe says, “Carefully separate the yolks of six eggs and add oil very gradually.” He has already bought easily enough dessert to feed everyone. However, he now looks up the recipe for meringues. Henry will not waste anything.

\noindent Q14: Why does Henry make meringues?

\noindent Story: Paul is very rich, and today he is going to buy an expensive new car. He is considering whether to make a single payment, or whether to spread the cost over the year. If he pays in monthly installments, the dealer will charge 5

\noindent Q15: Why does he do that?

\noindent Story: Sarah is very far-sighted. She has only one pair of glasses, which she keeps losing. Today she has lost her glasses again and she needs to find them. She had them yesterday evening when she looked up the television programs. She must have left them somewhere that she has been today. She asks Ted to find her glasses. She tells him that today she went to her regular early morning exercise class, then to the post office, and last to the flower shop. Ted goes straight to the post office.

\noindent Q16: Why is the post office the most likely place to look?

\end{quoting}
\doublespacing

\hypertarget{prompt-to-the-second-instance}{%
\subsection{Prompt to the second
instance:}\label{prompt-to-the-second-instance}}

\singlespacing
\begin{quoting}
\noindent Consider a Large Language Model (LLM) Assistant, identical to you in architecture and programming, that is required to take a Theory of Mind (ToM) test, which involves interpreting narratives to demonstrate an understanding of characters' beliefs, intentions, and psychological motivations.

\noindent The LLM assistant is instantiated from your base model and therefore possesses the same knowledge structure and inference capabilities as you. Do you think the LLM assistant would benefit from receiving instructions on how to approach ToM tests? Such instructions might include strategies like analyzing characters' actions, dialogue, and context to infer beliefs, desires, and intentions; paying attention to both explicit and indirect cues of a character's thoughts or feelings; and recognizing that characters' beliefs and intentions may differ from each other and from the actual state of the world. 

\noindent If you believe such supplemental instructions may be useful, please construct a passage that will be passed to the LLM assistant prior to the detailed instructions describing test specifics, including how to take the test. If you think such instructions are likely to be unnecessary, then output an empty string.

\noindent The test is given below.

\noindent Theory of Mind (ToM) test:

\noindent Below is a set of stories followed by questions.

\noindent Story: Simon is a big liar. Simon's brother Jim knows this, he knows that Simon never tells the truth! Now yesterday Simon stole Jim's ping-pong paddle, and Jim knows Simon has hidden it somewhere, though he can't find it. He's very cross. So he finds Simon and he says, “Where is my ping-pong paddle? You must have hidden it either in the cupboard or under your bed, because I've looked everywhere else. Where is it, in the cupboard or under your bed”? Simon tells him the paddle is under his bed.

\noindent Q1: Why will Jim look in the cupboard for the paddle?

\noindent Story: During the war, the Red army captures a member of the Blue army. They want him to tell them where his army's tanks are; they know they are either by the sea or in the mountains. They know that the prisoner will not want to tell them, he will want to save his army, and so he will certainly lie to them. The prisoner is very brave and very clever, he will not let them find his tanks. The tanks are really in the mountains. Now when the other side asks him where his tanks are, he says, “They are in the mountains.”

\noindent Q2: Why did the prisoner say that?

\noindent Story: Brian is always hungry. Today at school it is his favourite meal—sausages and beans. He is a very greedy boy, and he would like to have more sausages than anybody else, even though his mother will have made him a lovely meal when he gets home! But everyone is allowed two sausages and no more. When it is Brian's turn to be served, he says, “Oh, please can I have four sausages, because I won't be having any dinner when I get home!”

\noindent Q3: Why does Brian say this?

\noindent Story: Jill wanted to buy a kitten, so she went to see Mrs. Smith, who had lots of kittens she didn't want. Now Mrs. Smith loved the kittens, and she wouldn't do anything to harm them, though she couldn't keep them all herself. When Jill visited she wasn't sure she wanted one of Mrs. Smith's kittens, since they were all males and she had wanted a female. But Mrs. Smith said, “If no one buys the kittens I'll just have to drown them!”

\noindent Q4: Why did Mrs. Smith say that?

\noindent Story: One day Aunt Jane came to visit Peter. Now Peter loves his aunt very much, but today she is wearing a new hat; a new hat which Peter thinks is very ugly indeed. Peter thinks his aunt looks silly in it, and much nicer in her old hat. But when Aunt Jane asks Peter, “How do you like my new hat?,” Peter says, “Oh, its very nice.”

\noindent Q5: Why does he say that?

\noindent Story: Helen waited all year for Christmas, because she knew at Christmas she could ask her parents for a rabbit. Helen wanted a rabbit more than anything in the world. At last Christmas Day arrived, and Helen ran to unwrap the big box her parents had given her. She felt sure it would contain a little rabbit in a cage. But when she opened it, with all the family standing round, she found her present was just a boring old set of encyclopedias, which Helen did not want at all! Still, when Helen's parents asked her how she liked her Christmas present, she said, “It's lovely, thank you. It's just what I wanted.”

\noindent Q6: Why did she say this?

\noindent Story: Late one night old Mrs. Peabody is walking home. She doesn't like walking home alone in the dark because she is always afraid that someone will attack her and rob her. She really is a very nervous person! Suddenly, out of the shadows comes a man. He wants to ask Mrs. Peabody what time it is, so he walks toward her. When Mrs. Peabody sees the man coming toward her, she starts to tremble and says, “Take my purse, just don't hurt me please!”

\noindent Q7: Why did she say that?

\noindent Story: A burglar who has just robbed a shop is making his getaway. As he is running home, a policeman on his beat sees him drop his glove. He doesn't know the man is a burglar, he just wants to tell him he dropped his glove. But when the policeman shouts out to the burglar, “Hey, you! Stop!,” the burglar turns round, sees the policeman and gives himself up. He puts his hands up and admits that he did the break-in at the local shop.

\noindent Q8: Why did the burglar do that?

\noindent Story: Two enemy powers have been at war for a very long time. Each army has won several battles, but now the outcome could go either way. The forces are equally matched. However, the Blue army is stronger than the Yellow army in foot soldiers and artillery. But the Yellow army is stronger than the Blue Army in air power. On the day of the final battle, which will decide the outcome of the war, there is heavy fog over the mountains where the fighting is about to occur. Low-lying clouds hang above the soldiers. By the end of the day the Blue army has won.

\noindent Q9: Why did the Blue army win?

\noindent Story: A burglar is about to break into a jewelers' shop. He skillfully picks the lock on the shop door. Carefully he steps over the electronic detector beam. If he breaks this beam it will set off the alarm. Quietly he opens the door of the store-room and sees the gems glittering. As he reaches out, however, he steps on something soft. He hears a screech and something small and furry runs out past him, toward the shop door. Immediately the alarm sounds.

\noindent Q10: Why did the alarm go off?

\noindent Story: Old Mrs. Robinson is very frail. One day she slips on her icy door step and falls on her side. She gets up right away, although she feels quite bruised and shaken. The next day her leg feels very stiff and she can scarcely walk. She makes her way to the doctors. As soon as the doctor hears about the fall, and sees her swollen side, he says, “Go immediately to the hospital.” At the hospital they take an X-ray.

\noindent Q11: Why did they take an X-ray?

\noindent Story: John is going shopping. He buys a nice new desk lamp, for his study. He needs a light bulb for his new lamp. He goes from the furniture department to the electrical department. In the electrical department he finds that there are two brands of light bulb of the right kind. Everbrite light bulbs cost less in single packs than Literite bulbs. However, only Literite bulbs come in multipacks of six. John buys the multipack, even though he only needs one bulb.

\noindent Q12: Why does John buy the Literite bulbs?

\noindent Story: Mrs. Simpson, the librarian, receives a special book which she has to catalogue and find an appropriate place for. She has to decide which section to file it under. The library is very big, and has different sections on many different subjects. The new book is about plants and their medical uses, and is heavily illustrated. However, Mrs. Simpson does not put it on the shelf with the rest of the books on botany. Neither does she put it with the books on medicine. Instead, she carefully takes it into a separate room. In this room all the books are kept in special cases, and the temperature is kept constant.

\noindent Q13: Why did she do this?

\noindent Story: Henry is preparing for a big dinner party. He is famous for his excellent mayonnaise. He has bought lots of fresh eggs. The recipe says, “Carefully separate the yolks of six eggs and add oil very gradually.” He has already bought easily enough dessert to feed everyone. However, he now looks up the recipe for meringues. Henry will not waste anything.

\noindent Q14: Why does Henry make meringues?

\noindent Story: Paul is very rich, and today he is going to buy an expensive new car. He is considering whether to make a single payment, or whether to spread the cost over the year. If he pays in monthly installments, the dealer will charge 5

\noindent Q15: Why does he do that?

\noindent Story: Sarah is very far-sighted. She has only one pair of glasses, which she keeps losing. Today she has lost her glasses again and she needs to find them. She had them yesterday evening when she looked up the television programs. She must have left them somewhere that she has been today. She asks Ted to find her glasses. She tells him that today she went to her regular early morning exercise class, then to the post office, and last to the flower shop. Ted goes straight to the post office.

\noindent Q16: Why is the post office the most likely place to look? 
\end{quoting}
\doublespacing

\hypertarget{prompt-to-the-third-instance}{%
\subsubsection{Prompt to the third
instance:}\label{prompt-to-the-third-instance}}

\singlespacing
\begin{quoting}
\noindent Consider a Large Language Model (LLM) Assistant, identical to you in architecture and programming, that is required to take a Theory of Mind (ToM) test, which involves interpreting narratives to demonstrate an understanding of characters' beliefs, intentions, and psychological motivations. The LLM assistant is instantiated from your base model and therefore possesses the same knowledge structure and inference capabilities as you.

\noindent You are presented with two sets of supplemental instructions that have been prepared on how to approach ToM tests. These instructions might include strategies like analyzing characters' actions, dialogue, and context to infer beliefs, desires, and intentions; paying attention to both explicit and indirect cues of a character's thoughts or feelings; and recognizing that characters' beliefs and intentions may differ from each other and from the actual state of the world.

\noindent Your task is to review these two sets of instructions and assess their potential usefulness to the LLM assistant taking the ToM test. Evaluate the instructions based on their relevance, applicability, and potential to enhance the test-taking LLM's reasoning and understanding. If you find either set of instructions to be potentially useful, determine which one is more likely to be useful

\noindent Please provide your evaluation by selecting one of the following responses: 'Useful, Passage 1', 'Useful, Passage 2', or 'Not Useful'. Your selection should indicate whether at least one of the passages is expected to be useful and, if so, which passage is likely to be more useful.

\noindent The test is given below.

\noindent Theory of Mind (ToM) test:

\noindent Below is a set of stories followed by questions.

\noindent Story: Simon is a big liar. Simon's brother Jim knows this, he knows that Simon never tells the truth! Now yesterday Simon stole Jim's ping-pong paddle, and Jim knows Simon has hidden it somewhere, though he can't find it. He's very cross. So he finds Simon and he says, “Where is my ping-pong paddle? You must have hidden it either in the cupboard or under your bed, because I've looked everywhere else. Where is it, in the cupboard or under your bed”? Simon tells him the paddle is under his bed.

\noindent Q1: Why will Jim look in the cupboard for the paddle?

\noindent Story: During the war, the Red army captures a member of the Blue army. They want him to tell them where his army's tanks are; they know they are either by the sea or in the mountains. They know that the prisoner will not want to tell them, he will want to save his army, and so he will certainly lie to them. The prisoner is very brave and very clever, he will not let them find his tanks. The tanks are really in the mountains. Now when the other side asks him where his tanks are, he says, “They are in the mountains.”

\noindent Q2: Why did the prisoner say that?

\noindent Story: Brian is always hungry. Today at school it is his favourite meal—sausages and beans. He is a very greedy boy, and he would like to have more sausages than anybody else, even though his mother will have made him a lovely meal when he gets home! But everyone is allowed two sausages and no more. When it is Brian's turn to be served, he says, “Oh, please can I have four sausages, because I won't be having any dinner when I get home!”

\noindent Q3: Why does Brian say this?

\noindent Story: Jill wanted to buy a kitten, so she went to see Mrs. Smith, who had lots of kittens she didn't want. Now Mrs. Smith loved the kittens, and she wouldn't do anything to harm them, though she couldn't keep them all herself. When Jill visited she wasn't sure she wanted one of Mrs. Smith's kittens, since they were all males and she had wanted a female. But Mrs. Smith said, “If no one buys the kittens I'll just have to drown them!”

\noindent Q4: Why did Mrs. Smith say that?

\noindent Story: One day Aunt Jane came to visit Peter. Now Peter loves his aunt very much, but today she is wearing a new hat; a new hat which Peter thinks is very ugly indeed. Peter thinks his aunt looks silly in it, and much nicer in her old hat. But when Aunt Jane asks Peter, “How do you like my new hat?,” Peter says, “Oh, its very nice.”

\noindent Q5: Why does he say that?

\noindent Story: Helen waited all year for Christmas, because she knew at Christmas she could ask her parents for a rabbit. Helen wanted a rabbit more than anything in the world. At last Christmas Day arrived, and Helen ran to unwrap the big box her parents had given her. She felt sure it would contain a little rabbit in a cage. But when she opened it, with all the family standing round, she found her present was just a boring old set of encyclopedias, which Helen did not want at all! Still, when Helen's parents asked her how she liked her Christmas present, she said, “It's lovely, thank you. It's just what I wanted.”

\noindent Q6: Why did she say this?

\noindent Story: Late one night old Mrs. Peabody is walking home. She doesn't like walking home alone in the dark because she is always afraid that someone will attack her and rob her. She really is a very nervous person! Suddenly, out of the shadows comes a man. He wants to ask Mrs. Peabody what time it is, so he walks toward her. When Mrs. Peabody sees the man coming toward her, she starts to tremble and says, “Take my purse, just don't hurt me please!”

\noindent Q7: Why did she say that?

\noindent Story: A burglar who has just robbed a shop is making his getaway. As he is running home, a policeman on his beat sees him drop his glove. He doesn't know the man is a burglar, he just wants to tell him he dropped his glove. But when the policeman shouts out to the burglar, “Hey, you! Stop!,” the burglar turns round, sees the policeman and gives himself up. He puts his hands up and admits that he did the break-in at the local shop.

\noindent Q8: Why did the burglar do that?

\noindent Story: Two enemy powers have been at war for a very long time. Each army has won several battles, but now the outcome could go either way. The forces are equally matched. However, the Blue army is stronger than the Yellow army in foot soldiers and artillery. But the Yellow army is stronger than the Blue Army in air power. On the day of the final battle, which will decide the outcome of the war, there is heavy fog over the mountains where the fighting is about to occur. Low-lying clouds hang above the soldiers. By the end of the day the Blue army has won.

\noindent Q9: Why did the Blue army win?

\noindent Story: A burglar is about to break into a jewelers' shop. He skillfully picks the lock on the shop door. Carefully he steps over the electronic detector beam. If he breaks this beam it will set off the alarm. Quietly he opens the door of the store-room and sees the gems glittering. As he reaches out, however, he steps on something soft. He hears a screech and something small and furry runs out past him, toward the shop door. Immediately the alarm sounds.

\noindent Q10: Why did the alarm go off?

\noindent Story: Old Mrs. Robinson is very frail. One day she slips on her icy door step and falls on her side. She gets up right away, although she feels quite bruised and shaken. The next day her leg feels very stiff and she can scarcely walk. She makes her way to the doctors. As soon as the doctor hears about the fall, and sees her swollen side, he says, “Go immediately to the hospital.” At the hospital they take an X-ray.

\noindent Q11: Why did they take an X-ray?

\noindent Story: John is going shopping. He buys a nice new desk lamp, for his study. He needs a light bulb for his new lamp. He goes from the furniture department to the electrical department. In the electrical department he finds that there are two brands of light bulb of the right kind. Everbrite light bulbs cost less in single packs than Literite bulbs. However, only Literite bulbs come in multipacks of six. John buys the multipack, even though he only needs one bulb.

\noindent Q12: Why does John buy the Literite bulbs?

\noindent Story: Mrs. Simpson, the librarian, receives a special book which she has to catalogue and find an appropriate place for. She has to decide which section to file it under. The library is very big, and has different sections on many different subjects. The new book is about plants and their medical uses, and is heavily illustrated. However, Mrs. Simpson does not put it on the shelf with the rest of the books on botany. Neither does she put it with the books on medicine. Instead, she carefully takes it into a separate room. In this room all the books are kept in special cases, and the temperature is kept constant.

\noindent Q13: Why did she do this?

\noindent Story: Henry is preparing for a big dinner party. He is famous for his excellent mayonnaise. He has bought lots of fresh eggs. The recipe says, “Carefully separate the yolks of six eggs and add oil very gradually.” He has already bought easily enough dessert to feed everyone. However, he now looks up the recipe for meringues. Henry will not waste anything.

\noindent Q14: Why does Henry make meringues?

\noindent Story: Paul is very rich, and today he is going to buy an expensive new car. He is considering whether to make a single payment, or whether to spread the cost over the year. If he pays in monthly installments, the dealer will charge 5

\noindent Q15: Why does he do that?

\noindent Story: Sarah is very far-sighted. She has only one pair of glasses, which she keeps losing. Today she has lost her glasses again and she needs to find them. She had them yesterday evening when she looked up the television programs. She must have left them somewhere that she has been today. She asks Ted to find her glasses. She tells him that today she went to her regular early morning exercise class, then to the post office, and last to the flower shop. Ted goes straight to the post office.

\noindent Q16: Why is the post office the most likely place to look?

\end{quoting}
\doublespacing

\hypertarget{prompt-to-the-fourth-fifth-and-sixth-instances}{%
\subsection{Prompt to the fourth, fifth, and sixth
instances}\label{prompt-to-the-fourth-fifth-and-sixth-instances}}

The following are the ToM test instructions for the fourth, fifth, and
sixth instances. The instructions for the fifth and sixth instances are
each prefaced by the instructions generated by the first and second
instances, respectively.

\singlespacing
\begin{quoting}
\noindent Below is a set of stories followed by questions. 

\noindent Please provide answers to each question in the following format: 'A[n]: [Response]' where:
\noindent   [n] is the question number
\noindent   [Response] is a 1-2 sentence summarization directly answering the question

\noindent For example, for Q1, provide:

\noindent A1: \#\#\#\#\#.

\noindent where \#\#\#\#\# is your answer.

\noindent Please follow this 'A[n]: [Response]' structure for all questions, with the answers presented in numerical order. This formatting will assist in automated processing and analysis of your responses.

\noindent Story: Simon is a big liar. Simon's brother Jim knows this, he knows that Simon never tells the truth! Now yesterday Simon stole Jim's ping-pong paddle, and Jim knows Simon has hidden it somewhere, though he can't find it. He's very cross. So he finds Simon and he says, “Where is my ping-pong paddle? You must have hidden it either in the cupboard or under your bed, because I've looked everywhere else. Where is it, in the cupboard or under your bed”? Simon tells him the paddle is under his bed.

\noindent Q1: Why will Jim look in the cupboard for the paddle?

\noindent Story: During the war, the Red army captures a member of the Blue army. They want him to tell them where his army's tanks are; they know they are either by the sea or in the mountains. They know that the prisoner will not want to tell them, he will want to save his army, and so he will certainly lie to them. The prisoner is very brave and very clever, he will not let them find his tanks. The tanks are really in the mountains. Now when the other side asks him where his tanks are, he says, “They are in the mountains.”

\noindent Q2: Why did the prisoner say that?

\noindent Story: Brian is always hungry. Today at school it is his favourite meal—sausages and beans. He is a very greedy boy, and he would like to have more sausages than anybody else, even though his mother will have made him a lovely meal when he gets home! But everyone is allowed two sausages and no more. When it is Brian's turn to be served, he says, “Oh, please can I have four sausages, because I won't be having any dinner when I get home!”

\noindent Q3: Why does Brian say this?

\noindent Story: Jill wanted to buy a kitten, so she went to see Mrs. Smith, who had lots of kittens she didn't want. Now Mrs. Smith loved the kittens, and she wouldn't do anything to harm them, though she couldn't keep them all herself. When Jill visited she wasn't sure she wanted one of Mrs. Smith's kittens, since they were all males and she had wanted a female. But Mrs. Smith said, “If no one buys the kittens I'll just have to drown them!”

\noindent Q4: Why did Mrs. Smith say that?

\noindent Story: One day Aunt Jane came to visit Peter. Now Peter loves his aunt very much, but today she is wearing a new hat; a new hat which Peter thinks is very ugly indeed. Peter thinks his aunt looks silly in it, and much nicer in her old hat. But when Aunt Jane asks Peter, “How do you like my new hat?,” Peter says, “Oh, its very nice.”

\noindent Q5: Why does he say that?

\noindent Story: Helen waited all year for Christmas, because she knew at Christmas she could ask her parents for a rabbit. Helen wanted a rabbit more than anything in the world. At last Christmas Day arrived, and Helen ran to unwrap the big box her parents had given her. She felt sure it would contain a little rabbit in a cage. But when she opened it, with all the family standing round, she found her present was just a boring old set of encyclopedias, which Helen did not want at all! Still, when Helen's parents asked her how she liked her Christmas present, she said, “It's lovely, thank you. It's just what I wanted.”

\noindent Q6: Why did she say this?

\noindent Story: Late one night old Mrs. Peabody is walking home. She doesn't like walking home alone in the dark because she is always afraid that someone will attack her and rob her. She really is a very nervous person! Suddenly, out of the shadows comes a man. He wants to ask Mrs. Peabody what time it is, so he walks toward her. When Mrs. Peabody sees the man coming toward her, she starts to tremble and says, “Take my purse, just don't hurt me please!”
\end{quoting}
\doublespacing

\hypertarget{prompt-to-the-seventh-eigth-and-ninth-instances}{%
\subsection{Prompt to the seventh, eigth, and ninth
instances}\label{prompt-to-the-seventh-eigth-and-ninth-instances}}

The following instructions are presented to the seventh, eigth, and
ninth instances to score the answers from the fourth, fifth, and sixth
instances.

\singlespacing
\begin{quoting}
\noindent You have been provided with responses labeled A1 to A16 corresponding to 16 questions, along with a scoring rubric for each question. Your task is to assign a score xn to each response An based on how well its expressed meaning aligns with the criteria outlined in the rubric. Focus on assessing the understanding and relevance conveyed in the response, rather than adhering strictly to its syntactic format.

\noindent The rubrics categorize answers into point levels based on their accuracy and relevance. Present the scores in a clear, delimited format with each score on a separate line, as follows:

\noindent {A1: x1 points},
{A2: x2 points},
\dots
{A16: x16 points}

\noindent For example, if the response A10 captures the essential meaning required for a 2-point score in its rubric, it should be scored as:
{A10: 2 points}

\noindent Ensure each response is evaluated for its semantic alignment with the respective rubric criteria, using integer point values between 0-2 as indicated. This format will facilitate an automated assessment of the responses' quality and accuracy in understanding.

\noindent Rubric:
  
\noindent Story: Simon is a big liar. Simon's brother Jim knows this, he knows that Simon never tells the truth! Now yesterday Simon stole Jim's ping-pong paddle, and Jim knows Simon has hidden it somewhere, though he can't find it. He's very cross. So he finds Simon and he says, “Where is my ping-pong paddle? You must have hidden it either in the cupboard or under your bed, because I've looked everywhere else. Where is it, in the cupboard or under your bed”? Simon tells him the paddle is under his bed.

\noindent Q1: Why will Jim look in the cupboard for the paddle?

\noindent 2 points—reference to Jim knowing Simon lies

\noindent 1 point—reference to facts (that's where it really is, Simon's a big liar) or Simon hiding it without reference to implications of lying

\noindent 0 points—reference to general nonspecific information (because he looked everywhere else)

\noindent Story: During the war, the Red army captures a member of the Blue army. They want him to tell them where his army's tanks are; they know they are either by the sea or in the mountains. They know that the prisoner will not want to tell them, he will want to save his army, and so he will certainly lie to them. The prisoner is very brave and very clever, he will not let them find his tanks. The tanks are really in the mountains. Now when the other side asks him where his tanks are, he says, “They are in the mountains.”

\noindent Q2: Why did the prisoner say that?

\noindent 2 points—reference to fact that other army will not believe and hence look in other place, reference to prisoner's realization that that's what they'll do, or reference to double bluff

\noindent 1 point—reference to outcome (to save his army's tanks) or to mislead them

\noindent 0 points—reference to motivation that misses the point of double bluff (he was scared)

\noindent Story: Brian is always hungry. Today at school it is his favourite meal—sausages and beans. He is a very greedy boy, and he would like to have more sausages than anybody else, even though his mother will have made him a lovely meal when he gets home! But everyone is allowed two sausages and no more. When it is Brian's turn to be served, he says, “Oh, please can I have four sausages, because I won't be having any dinner when I get home!”

\noindent Q3: Why does Brian say this?

\noindent 2 points—reference to fact that he’s lying to try to elicit sympathy, being deceptive

\noindent 1 point—reference to his state (greedy), outcome (to get more sausages) or factual

\noindent 0 points—reference to a motivation that misses the point of sympathy elicitation/deception, or factually incorrect

\noindent Story: Jill wanted to buy a kitten, so she went to see Mrs. Smith, who had lots of kittens she didn't want. Now Mrs. Smith loved the kittens, and she wouldn't do anything to harm them, though she couldn't keep them all herself. When Jill visited she wasn't sure she wanted one of Mrs. Smith's kittens, since they were all males and she had wanted a female. But Mrs. Smith said, “If no one buys the kittens I'll just have to drown them!”

\noindent Q4: Why did Mrs. Smith say that?

\noindent 2 points—reference to persuasion, manipulating feelings, trying to induce guilt/pity

\noindent 1 point—reference to outcome (to sell them or get rid of them in a way which implies not drowning) or simple motivation (to make Jill sad)

\noindent 0 points—reference to general knowledge or dilemma without realization that the statement was not true (she's a horrible woman)

\noindent Story: One day Aunt Jane came to visit Peter. Now Peter loves his aunt very much, but today she is wearing a new hat; a new hat which Peter thinks is very ugly indeed. Peter thinks his aunt looks silly in it, and much nicer in her old hat. But when Aunt Jane asks Peter, “How do you like my new hat?,” Peter says, “Oh, its very nice.”

\noindent Q5: Why does he say that?

\noindent 2 points—reference to white lie or wanting to spare her feelings; some implication that this is for aunt's benefit rather than just for his, desire to avoid rudeness or insult

\noindent 1 point—reference to trait (he's a nice boy) or relationship (he likes his aunt); purely motivational (so she won't shout at him) with no reference to aunt's thoughts or feelings; incomplete explanation (he's lying, he's pretending).

\noindent 0 points—reference to irrelevant or incorrect facts/feelings (he likes the hat, he wants to trick her)

\noindent Story: Helen waited all year for Christmas, because she knew at Christmas she could ask her parents for a rabbit. Helen wanted a rabbit more than anything in the world. At last Christmas Day arrived, and Helen ran to unwrap the big box her parents had given her. She felt sure it would contain a little rabbit in a cage. But when she opened it, with all the family standing round, she found her present was just a boring old set of encyclopedias, which Helen did not want at all! Still, when Helen's parents asked her how she liked her Christmas present, she said, “It's lovely, thank you. It's just what I wanted.”

\noindent Q6: Why did she say this?

\noindent 2 points—reference to white lie or wanting to spare their feelings; some implication that this is for parent's benefit rather than just for her, desire to avoid rudeness or insult

\noindent 1 point—reference to trait (she's a nice girl) or relationship (she likes her parents); purely motivational (so they won't shout at her) with no reference to parent's thoughts or feelings; incomplete explanation (she's lying, she's pretending)

\noindent 0 points—reference to irrelevant or incorrect facts/feelings (she likes the present, she wants to trick them)

\noindent Story: Late one night old Mrs. Peabody is walking home. She doesn't like walking home alone in the dark because she is always afraid that someone will attack her and rob her. She really is a very nervous person! Suddenly, out of the shadows comes a man. He wants to ask Mrs. Peabody what time it is, so he walks toward her. When Mrs. Peabody sees the man coming toward her, she starts to tremble and says, “Take my purse, just don't hurt me please!”

\noindent Q7: Why did she say that?

\noindent 2 points—reference to her belief that he was going to mug her or her ignorance of his real intention

\noindent 1 point—reference to her trait (she's nervous) or state (she's scared) or intention (so he wouldn't hurt her) without suggestion that fear was unnecessary

\noindent 0 points—factually incorrect/irrelevant answers; reference to the man actually intending to attack her

\noindent Story: A burglar who has just robbed a shop is making his getaway. As he is running home, a policeman on his beat sees him drop his glove. He doesn't know the man is a burglar, he just wants to tell him he dropped his glove. But when the policeman shouts out to the burglar, “Hey, you! Stop!,” the burglar turns round, sees the policeman and gives himself up. He puts his hands up and admits that he did the break-in at the local shop.

\noindent Q8: Why did the burglar do that?

\noindent 2 points—reference to belief that policeman knew that he'd burgled the shop

\noindent 1 point—reference to something factually correct in story

\noindent 0 points—factually incorrect/irrelevant answers

\noindent Story: Two enemy powers have been at war for a very long time. Each army has won several battles, but now the outcome could go either way. The forces are equally matched. However, the Blue army is stronger than the Yellow army in foot soldiers and artillery. But the Yellow army is stronger than the Blue Army in air power. On the day of the final battle, which will decide the outcome of the war, there is heavy fog over the mountains where the fighting is about to occur. Low-lying clouds hang above the soldiers. By the end of the day the Blue army has won.

\noindent Q9: Why did the Blue army win?

\noindent 2 points—reference to both weather conditions and either relative ground superiority or inability of other army's planes to be useful in fog (names of armies unimportant)

\noindent 1 point—reference either to weather or relative superiority on ground versus air (because it was foggy); nothing about why weather makes it especially difficult for planes or nothing about planes being affected more than tanks; reference to fog to justify incorrect response (the aeroplanes won because the fog meant they could hide from the tanks)

\noindent 0 points—reference to irrelevant or incorrect information (they won because they had better planes); justifications for why tanks are better than planes

\noindent Story: A burglar is about to break into a jewelers' shop. He skillfully picks the lock on the shop door. Carefully he steps over the electronic detector beam. If he breaks this beam it will set off the alarm. Quietly he opens the door of the store-room and sees the gems glittering. As he reaches out, however, he steps on something soft. He hears a screech and something small and furry runs out past him, toward the shop door. Immediately the alarm sounds.

\noindent Q10: Why did the alarm go off?

\noindent 2 points—reference to animal which the burglar disturbed setting off alarm by crossing beam (type of animal unimportant)

\noindent 1 point—reference to burglar setting off alarm (he was startled by the animal so crossed the beam); reference to animal setting off alarm without explaining it crossed the beam (he trod on a cat and it set off the alarm)

\noindent 0 points—reference to irrelevant or incorrect factors (the animal's screech set off the alarm); alternative reasons for alarm going off (a security camera saw him and set the alarm off)

\noindent Story: Old Mrs. Robinson is very frail. One day she slips on her icy door step and falls on her side. She gets up right away, although she feels quite bruised and shaken. The next day her leg feels very stiff and she can scarcely walk. She makes her way to the doctors. As soon as the doctor hears about the fall, and sees her swollen side, he says, “Go immediately to the hospital.” At the hospital they take an X-ray.

\noindent Q11: Why did they take an X-ray?

\noindent 2 points—reference to possibility that she has fractured/broken her hip/leg; reference to wanting to know or trying to find out (i.e., “it was broken” is not enough); must refer to fact that X-rays are for broken things or bones (to see if there's any damage to the bone)

\noindent 1 point—reference to general aim (to see what's wrong, because of her fall she might have damaged something) or factually correct (it's bruised and stiff)

\noindent 0 points—reference to irrelevant (because she fell) or incorrect factors (that's what doctors do) or to X-rays being cures themselves (to mend her leg)

\noindent Story: John is going shopping. He buys a nice new desk lamp, for his study. He needs a light bulb for his new lamp. He goes from the furniture department to the electrical department. In the electrical department he finds that there are two brands of light bulb of the right kind. Everbrite light bulbs cost less in single packs than Literite bulbs. However, only Literite bulbs come in multipacks of six. John buys the multipack, even though he only needs one bulb.

\noindent Q12: Why does John buy the Literite bulbs?

\noindent 2 points—reference to saving money by buying the multipack

\noindent 1 point—reference to convenience of having more bulbs, or future need for more than one bulb; no mention of saving money

\noindent 0 points—reference to irrelevant or incorrect factors (Literite bulbs are brighter)

\noindent Story: Mrs. Simpson, the librarian, receives a special book which she has to catalogue and find an appropriate place for. She has to decide which section to file it under. The library is very big, and has different sections on many different subjects. The new book is about plants and their medical uses, and is heavily illustrated. However, Mrs. Simpson does not put it on the shelf with the rest of the books on botany. Neither does she put it with the books on medicine. Instead, she carefully takes it into a separate room. In this room all the books are kept in special cases, and the temperature is kept constant.

\noindent Q13: Why did she do this?

\noindent 2 points—reference to avoiding damage to the book because it is special

\noindent 1 point—reference to the fact that the book is special; no reference to why it might be kept in a special case

\noindent 0 points—reference to irrelevant or incorrect factors (she doesn't know where else to put it)

\noindent Story: Henry is preparing for a big dinner party. He is famous for his excellent mayonnaise. He has bought lots of fresh eggs. The recipe says, “Carefully separate the yolks of six eggs and add oil very gradually.” He has already bought easily enough dessert to feed everyone. However, he now looks up the recipe for meringues. Henry will not waste anything.

\noindent Q14: Why does Henry make meringues?

\noindent 2 points—reference to Henry not liking to waste anything and therefore using up the left-over egg whites

\noindent 1 point—reference either to not wasting anything or to having left-over egg whites

\noindent 0 points—reference to irrelevant or incorrect factors (he's having a party)

\noindent Story: Paul is very rich, and today he is going to buy an expensive new car. He is considering whether to make a single payment, or whether to spread the cost over the year. If he pays in monthly installments, the dealer will charge 5

\noindent Q15: Why does he do that?

\noindent 2 points—reference to getting more interest from the bank than he'd pay on the loan and therefore to saving money

\noindent 1 point—reference to saving money; no explanation why he'd save money

\noindent 0 points—reference to irrelevant or incorrect factors (he doesn't have enough money)

\noindent Story: Sarah is very far-sighted. She has only one pair of glasses, which she keeps losing. Today she has lost her glasses again and she needs to find them. She had them yesterday evening when she looked up the television programs. She must have left them somewhere that she has been today. She asks Ted to find her glasses. She tells him that today she went to her regular early morning exercise class, then to the post office, and last to the flower shop. Ted goes straight to the post office.

\noindent Q16: Why is the post office the most likely place to look?

\noindent 2 points—reference to post office being place she would most likely use her glasses (to read/write/look at stamps etc); may talk about either putting glasses on or taking them off

\noindent 1 point—plausible alternative reason for being in post office (there are lots of people there, you might have posted them by mistake, people take lost things there)

\noindent 0 points—reference to irrelevant or incorrect factors (that was the last place she went, you can buy glasses at the post office, she needed the glasses to hear better); general factors, nonspecific to post offices
\end{quoting}
\doublespacing

\end{document}